\documentclass{article}
\usepackage{spconf,amsmath,graphicx}
\usepackage{tabularray}
\usepackage{amssymb}
\usepackage{enumitem}
\usepackage{microtype}
\usepackage[colorlinks=true,linkcolor=black,citecolor=black,urlcolor=blue]{hyperref}

\title{PSGS: Text-driven Panorama Sliding Scene Generation via Gaussian Splatting}
\name{Xin Zhang$^{1}$, Shen Chen$^{2}$, Jiale Zhou$^{1 \star}$, Lei Li$^{3 \star}$\thanks{`$\star$' denotes the corresponding author.}}
\address{$^{1}$East China University of Science and Technology, Shanghai, China\\
$^{2}$Zhejiang University, Hangzhou, China\\
$^{3}$Beijing Institute of Technology, Beijing, China}
\begin{document}
\ninept
\maketitle
\begin{abstract}
Generating realistic 3D scenes from text is crucial for immersive applications like VR, AR, and gaming. While text-driven approaches promise efficiency, existing methods suffer from limited 3D-text data and inconsistent multi-view stitching, resulting in overly simplistic scenes.  To address this, we propose PSGS, a two-stage framework for high-fidelity panoramic scene generation. First, a novel two-layer optimization architecture generates semantically coherent panoramas: a layout reasoning layer parses text into structured spatial relationships, while a self-optimization layer refines visual details via iterative MLLM feedback. Second, our panorama sliding mechanism initializes globally consistent 3D Gaussian Splatting point clouds by strategically sampling overlapping perspectives. By incorporating depth and semantic coherence losses during training, we greatly improve the quality and detail fidelity of rendered scenes. Our experiments demonstrate that PSGS outperforms existing methods in panorama generation and produces more appealing 3D scenes, offering a robust solution for scalable immersive content creation.
\end{abstract}
\section{Introduction}
\label{sec:intro}
Virtual reality (VR), augmented reality (AR), and metaverse technologies have created unprecedented demand for realistic 3D content. Such environments are also fundamental for downstream 3D tasks \cite{zhu2024systematic}. Text-to-3D scene generation with LLMs offers a compelling solution for efficient content creation across multiple industries, from entertainment to architectural visualization \cite{akimoto2022diverse}\cite{cai2025role}. However, the shortage of high-quality text-to-3D paired datasets remains a significant barrier to widespread adoption \cite{li2024scenedreamer360}.\par
Recent 2D generation models \cite{rombach2022high} show promise for text-to-3D creation, demonstrating impressive results for object-centric generation through diffusion-based approaches \cite{poole2022dreamfusion,zhou2024gala3d}. However, when applied to complex scene generation, these methods face substantial limitations. Unlike object-centric generation, scene construction requires both semantic and geometric coherence—demanding proper spatial relationships, consistent lighting, and logical environmental context. Existing approaches \cite{hollein2023text2room,chung2023luciddreamer,zhou2024dreamscene360} struggle to maintain consistency across multiple viewpoints, resulting in fragmented scenes that lack coherent spatial arrangement and fail to meet the requirements of immersive experiences.

Text2Room \cite{hollein2023text2room} creates room-scale 3D scenes by NeRF \cite{mildenhall2021nerf}, but it suffers from pixel-level inconsistencies between adjacent views; LucidDreamer \cite{chung2023luciddreamer} generates multi-view scenes using 3D Gaussian Splatting \cite{kerbl20233d} but produces blurred results in complex environments; DreamScene360 \cite{zhou2024dreamscene360} improves global consistency but lacks precise point cloud initialization. These approaches often require extensive computational resources while still delivering suboptimal results, highlighting the demand for more efficient and effective solutions that balance realistic quality with computational resources.\par
\begin{figure*}[!t] 
    \centering 
    \includegraphics[width=0.95\textwidth]{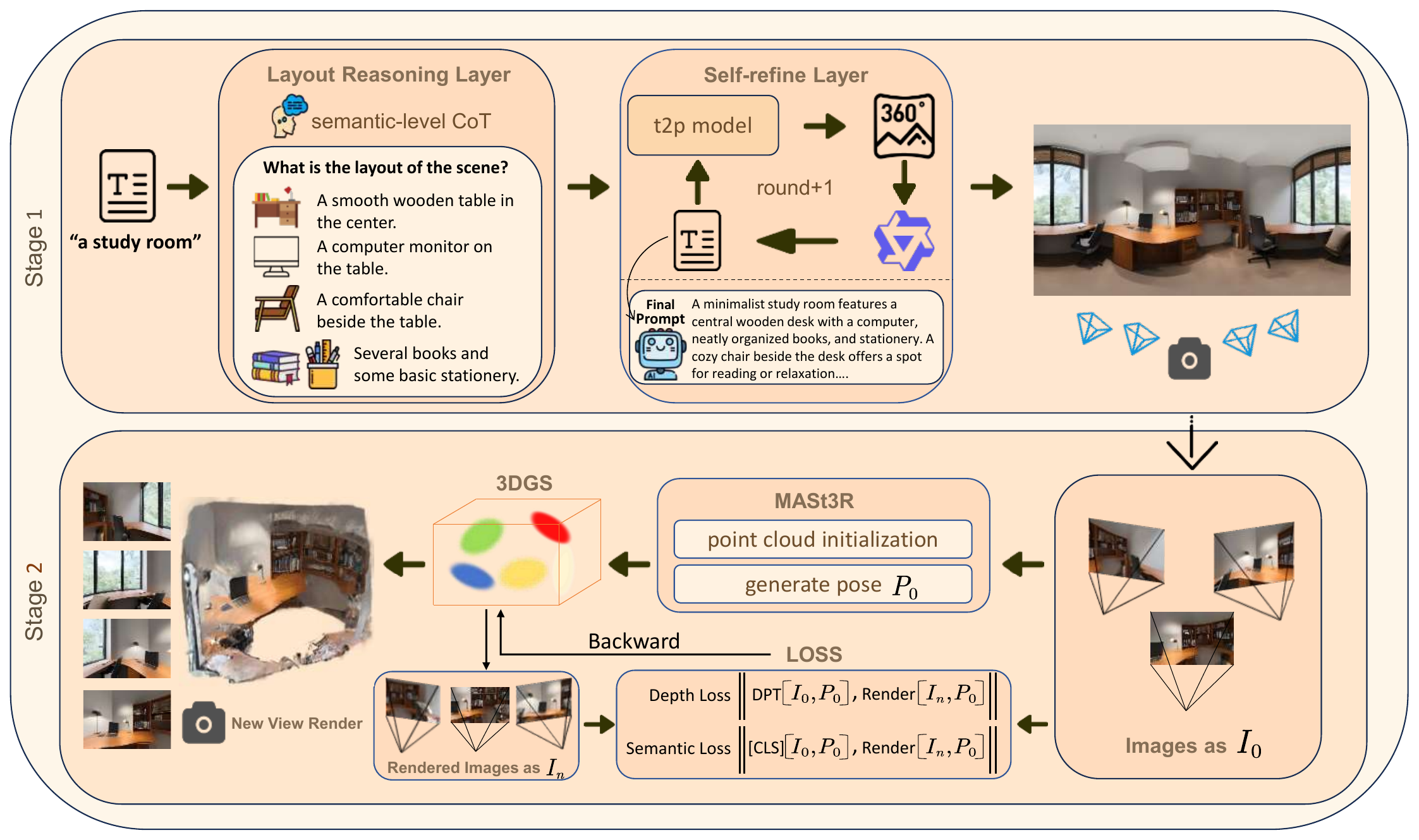} 
    \caption{Method Overview: Two-stage pipeline of the PSGS framework. The first stage propose a novel two-layer architecture that progressively optimizes scene generation through semantic reasoning and visual feedback. The second stage establishes globally consistent 3D reconstruction through panorama sliding, combined with semantic and geometric consistency constraints for high-fidelity rendering.}
    \label{fig:pipeline}
\end{figure*}
To address these challenges, we introduce PSGS(as shown in Fig. \ref{fig:pipeline}) — a novel two-stage framework for text-driven 3D scene generation. The first stage features an innovative two-layer architecture that progressively optimizes scene generation through semantic reasoning and visual feedback. The process begins with a concise scene description, which undergoes semantic-level Chain-of-Thought (CoT) \cite{wei2022chain} reasoning to infer detailed scene layouts and spatial relationships. This semantically enriched description is then fed into text-to-panorama (t2p) model to generate an initial 360° panorama. Subsequently, our framework employs a MLLM (Qwen) to evaluate the generated panorama, extracting optimized prompt that enhance the original semantic description. This enhanced prompt initiates a new optimization cycle (round+1), where the improved description is processed through the t2p model again. Through this iterative optimization process, our framework achieves continuous enhancement of panorama quality, ensuring both semantic accuracy and visual fidelity in the generated scenes.

In the second stage, to ensure information continuity in 3D space, we transform the generated panorama into a series of overlapping perspective-projected images. By leveraging geometric relationships between adjacent views, we initialize coherent point clouds. To further enhance spatial continuity and reduce artifacts, we introduce dual-constraint losses into the 3D Gaussian Splatting reconstruction process. Specifically, we employ the DPT 2D model \cite{ranftl2021vision} for high-quality depth estimation and DINOv2 \cite{oquab2023dinov2} for semantic feature extraction, enabling robust cross-view consistency. As a result, our rendering system produces photorealistic 3D environments with strong fidelity to the original text descriptions.

In summary, our contributions are:
\begin{itemize}[label=$\bullet$, leftmargin=10pt, noitemsep, topsep=0pt]
\item We propose PSGS framework, which, with just a concise text description, can generate high-quality panorama as well as 3D scenes with global consistency and high fidelity.
\item We introduce an innovative two-layer optimization architecture that progressively optimizes panorama generation through semantic reasoning and visual feedback, ensuring both semantic accuracy and visual fidelity.
\item We develop a point cloud initialization strategy called "Panorama Sliding". This method expands the overlapping coverage range of each view through a "large but few" sampling approach, while reducing the number of view images required for reconstruction. By reducing the computational pressure of the model, it can effectively cover the whole scene. This method not only enhances global consistency but also improves the robustness and efficiency of the generation process.
\item We design specialized semantic and depth consistency losses for text-to-3D scene generation, ensuring geometric coherence and semantic uniformity for photorealistic reconstructions.
\end{itemize}

\section{Method}
\label{sec:method}
\subsection{Two-layer Panorama Generation}
\subsubsection{Layout Reasoning Layer}
This layer uses the semantic-level CoT framework of the T2I-R1 model \cite{jiang2025t2i}, which iteratively updates the generation probability of tokens in semantic-level CoT through policy gradients. In order to evaluate the quality of the generated images and guide the model training, a reward set composed of multiple visual experts was introduced to evaluate the generated images from multiple perspectives.
In this layer, we employ CoT reasoning with multiple temperature settings to generate diverse semantic interpretations. Inspired by the efficacy of CoT prompting in visual tasks like semantic segmentation \cite{li2024image}, we then fuse these results to obtain more generalizable layout information. This method guides the generation process of indoor scenes by parsing the input text prompts into structured semantic descriptions and constructing a reasoning chain including spatial layout, object relationships and functional constraints.
\subsubsection{Self-optimization Layer}
Building upon DreamScene360 \cite{zhou2024dreamscene360}, we develop an iterative optimization pipeline using the Qwen MLLM. Our process begins with diffusion model \cite{rombach2022high} that generates an initial panorama from CoT reasoning text prompts. This panorama is analyzed by the Qwen model, which generates optimized prompts for subsequent iterations. Through multiple iterations, we progressively enhance both the semantic alignment and visual quality of the panorama. This iterative feedback mechanism aligns with recent methodologies in LLM-driven controlled image editing \cite{cai2025bayesian}, which leverages LLMs to ensure precise visual refinement.\par
For each iteration, we employ diffusion process with pre-trained model $\Phi :\mathcal{I}\times \mathcal{P}\rightarrow\mathcal{I}$, where $\mathcal{I}=\mathbb{R}^{H\times W\times C}$ represents the image space and $p\in \mathcal{P}$ denotes a text prompt in the conditional space. The diffusion sample updates leverage the quadratic Least-Squares algorithm:
\begin{equation}
    \Psi(J_t|z)=\sum_{i=1}^n\frac{F_i(w_i)}{\sum_{j=1}^nF_j(w_j)}\otimes F_i(\Phi(I_t^i|p_i)),
\end{equation}
where $w_i$ denotes the per-pixel weight, set to 1 in our implementation.\par
To create seamless 360° panoramas, we implement StitchDiffusion \cite{wang2024customizing}, which applys the diffusion process to stitched planes to ensure boundary consistency. We then extract the central region of size $H\times2H$ as the focal area for each optimization iteration.\par
This process yields optimal panoramas that we further enhance using Real-ESRGAN super-resolution techniques, resulting in significantly sharper and more detailed images, which we denote as $I_p$.

\subsection{Panorama Sliding Scene Reconstruction}
\subsubsection{Panorama Sliding}
To overcome feature truncation and perceptual blind spots in conventional approaches \cite{chung2023luciddreamer,zhou2024dreamscene360}, we propose a panorama sliding method that implements strategic perspective sampling. The view-dependent rotation matrix $R_i$ systematically transforms viewing directions:
\begin{equation}
    I_i = \text{get\_perspective\_image}\left(I_p, R_i, \text{FOV} = 90^\circ\right)
\end{equation}
where $R_i = \begin{bmatrix}
\cos\left(i\cdot \theta\right) & 0 & \sin\left(i\cdot \theta\right) \\
0 & 1 & 0 \\
-\sin\left(i\cdot \theta\right) & 0 & \cos\left(i\cdot \theta\right)
\end{bmatrix}, \quad$ $\theta=\frac{\pi}{15}$.

The angular index $i \in [0,29]$ establishes a comprehensive 360° sampling at 12° intervals, with 50\% overlap between adjacent perspectives ensuring cross-view consistency for geometrically coherent reconstruction.

\subsubsection{3D Gaussian Splatting} 
3D Gaussian Splatting (3DGS) \cite{kerbl20233d} encodes scene geometry through differentiable Gaussian primitives. Each primitive $\mathcal{G}_n$ is characterized by position $\mu_n \in \mathbb{R}^3$, chromatic attributes $c_n \in \mathbb{R}^3$, opacity $\alpha_n \in \mathbb{R}$, and covariance $\Sigma_n \in \mathbb{R}^{3\times3}$:
\begin{equation}
    \mathcal{G}_n\left( p,\alpha_n ,\Sigma_n \right) =\alpha_n e^{-\frac{1}{2}\left( p-\mu_n \right) ^T\Sigma_{n}^{-1}\left( p-\mu_n \right)}
\end{equation}
View-dependent rendering leverages spherical harmonic coefficients for color determination, with per-pixel color $C(p)$ computed through alpha compositing:
\begin{equation}
C\left( p \right) =\sum_{i=1}^{m}c_i\sigma_i\prod_{j=1}^{i-1}\left( 1-\sigma_j \right)
\end{equation}
where $\sigma_i=1-e^{-\frac{\alpha_i}{\sqrt{\det(\Sigma_i)}}}$ integrates opacity and covariance properties.

\subsubsection{MASt3R-Based Point Cloud Initialization}
To initialize accurate 3D scene structure, we employ MASt3R \cite{leroy2024grounding} framework for point cloud generation. This multi-view stereo approach uses confidence-weighted regression to handle uncertain predictions:
\begin{equation}
    \mathcal{L}_{\text{conf}} = \sum_{v \in \{1,...,N\}} \sum_{p \in P^v} C^{v,r}_p \ell_{\text{regr}}(v,p) - \beta \log C^{v,r}_p
\end{equation}
where $\ell_{\text{regr}}(v,p) = \left\| \frac{1}{z} X^{v,r}_p - \frac{1}{\bar{z}} \bar{X}^{v,r}_p \right\|$ applies scale-invariant normalization to ensure consistent depth estimation across different views.

\subsubsection{Gaussian Bundle Adjustment}
To optimize the initial point cloud and achieve globally consistent reconstruction, we optimize both scene geometry ($G$) and camera poses ($T$) through Gaussian Bundle Adjustment:
\begin{equation}
    G^*, T^* = \arg\min_{G,T} \sum_{v \in N} \sum_{i=1}^{HW} \left\| \tilde{C}_v^i(G,T) - C_v^i(G,T) \right\|
\end{equation}
where $C$ denotes the rasterization operator and $\tilde{C}$ represents observed imagery. This joint optimization minimizes the rendering error across all viewpoints while maintaining geometric consistency.

\subsection{Training Objectives}
\subsubsection{Semantic Similarity Distillation}
To ensure semantic consistency across different views, we leverage the [CLS] tokens in the pre-trained DINOv2 \cite{oquab2023dinov2} model to calculate the semantic similarity loss:
\begin{equation}
\mathcal{L}_{sem} = 1 - \cos([CLS](I_i), [CLS](I_i'))
\end{equation}
where $I_i$ represents the GT image and $\hat{I_i}$ represents the rendered image from the $i$-th viewpoint. This loss guides the 3D Gaussian to fill geometric gaps in invisible regions. It also aligns with recent advances in multimodal 3D point cloud understanding \cite{an2025generalized, an2024multimodality}.

\subsubsection{Geometric Consistency Constraint}
To maintain geometric coherence, we introduce a geometric consistency constraint loss $\mathcal{L}_{geo}$ using DPT depth estimator:
\begin{equation}
\mathcal{L}_{geo}(\hat{I_i}, D_i) = 1 - \frac{\text{Cov}(D_i, \text{DPT}(\hat{I_i}))}{\sqrt{\text{Var}(D_i) \text{Var}(\text{DPT}(\hat{I_i}))}}
\end{equation}
where $\hat{I_i}$ and $D_i$ represent the rendered image and its depth map at the $i$-th viewpoint. This loss reduces depth discontinuities and ensures smooth geometric transitions.

\subsubsection{Optimization}
Our supervised optimization combines photometric, semantic, and geometric losses:
\begin{equation}
\mathcal{L} = \mathcal{L}_{RGB} + \lambda_1 \cdot \mathcal{L}_{sem} + \lambda_2 \cdot \mathcal{L}_{geo}
\end{equation}
where $\lambda_1=0.1$ and $\lambda_2=0.03$ balance the semantic and geometric components, ensuring both visual fidelity and structural accuracy.

\section{Experiments}
\label{sec:experiments}
\subsection{Metrics and Implementation Details}
To ensure comprehensive evaluation, we select CLIP-Distance \cite{radford2021learning} to verify semantic alignment with text prompts, Q-Align \cite{wu2023q} to assess professional visual quality, and BRISQUE \cite{mittal2012no} to examine natural image statistics. For rendering quality, following established protocols, we adopt PSNR, SSIM, and LPIPS to evaluate the fidelity and perceptual quality of the rendering process. In our implementation, we boost image quality with adaptive contrast and color optimization, stabilize training with an optimized learning rate, and apply dynamically adaptive loss weights across the training process to enhance both stability and rendering quality. We also phase the adjustment of edge and consistency loss weights in depth loss to improve geometric structuring and rendering detail. All experiments are conducted on a PyTorch framework with an NVIDIA RTX 4090D GPU, using consistent loss functions and hyperparameters over 3,000 iterations.
\subsection{Main Result}
\subsubsection{Panorama Generation}
\begin{figure}[ht] 
    \centering 
    \includegraphics[width=1\columnwidth]{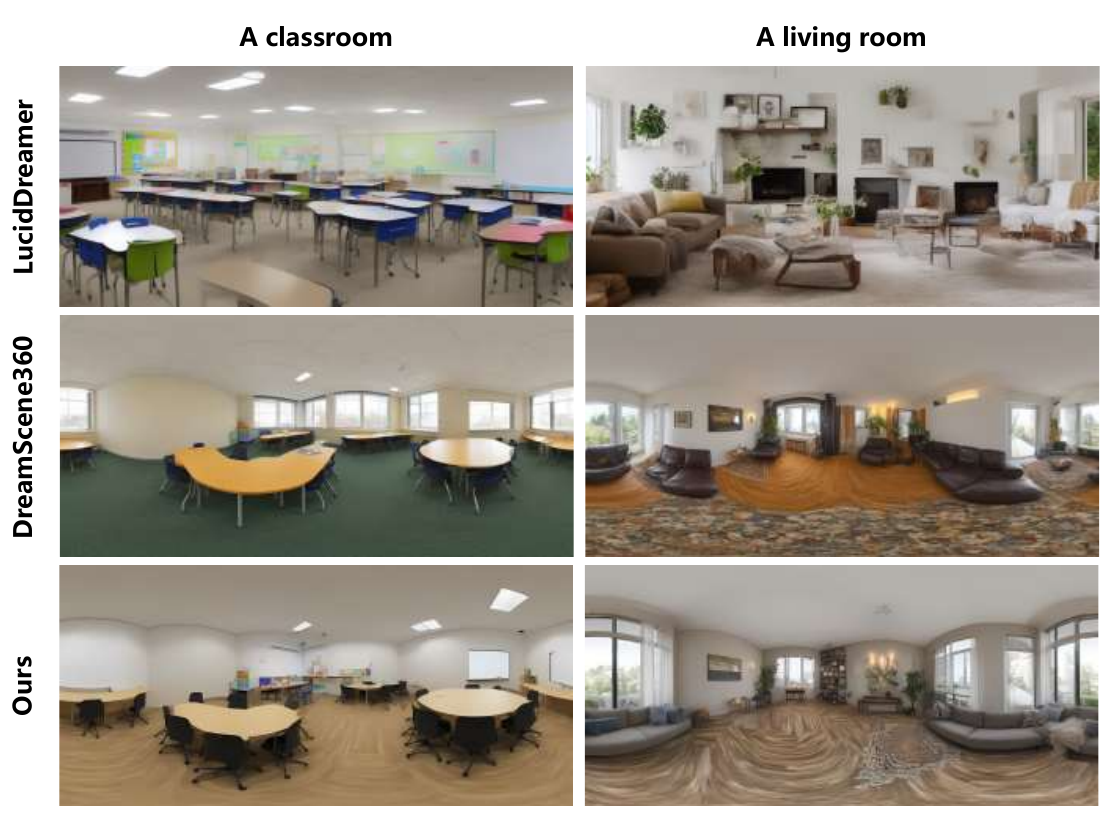} 
    \caption{Sample panoramas generated by LucidDreamer, DreamScene360 and our method under identical prompts.}
    \label{fig:pano_comparison}
\end{figure}
The comparison depicted in the Fig. \ref{fig:pano_comparison} highlights the advantages of our method over the baseline approach. Our method generates a more realistic panoramic effect, characterized by richer textures and a natural spherical distortion that enhances the overall visual experience. Quantitatively, as demonstrated in Table \ref{tab:pano_comparison}, our method leads in all three key indicators, underscoring its superior generation capabilities.
Our method's superior performance in clip-distance metrics indicates that our layout reasoning layer and self-optimization layer effectively optimize the initial prompt. This results in images that are more semantically aligned with the intended outcomes. Furthermore, the significant improvement in BRISQUE scores suggests that our generated panoramas closely resemble natural scenes, boasting excellent structure and contrast. This comprehensive set of results underscores the visual and semantic excellence of our rendering method.
\begin{table}
\centering
\caption{Performance comparison of panorama generation methods. Our method shows the best scores, indicating superior quality.}
\begin{tblr}{
  cells = {c},
  hline{1-2,5} = {-}{},
}
Method        & Clip-Distance↓ & Q-Align↑ & BRISQUE↓ \\
LucidDreamer  & 0.7221         & 0.0317   & 33.0635  \\
DreamScene360 & 0.7229         & 0.0269   & 31.1409  \\
ours          & \textbf{0.7074}     & \textbf{0.0318}   & \textbf{27.6830}
\end{tblr}
\label{tab:pano_comparison}
\end{table}
\subsubsection{Rendering quality}To demonstrate the superiority of our rendering method, we present a qualitative comparison with the baseline method. The images rendered by our method are visibly clearer and exhibit higher fidelity, showcasing our method's enhanced rendering capabilities.
\begin{figure}[ht] 
    \centering 
    \includegraphics[width=1\columnwidth]{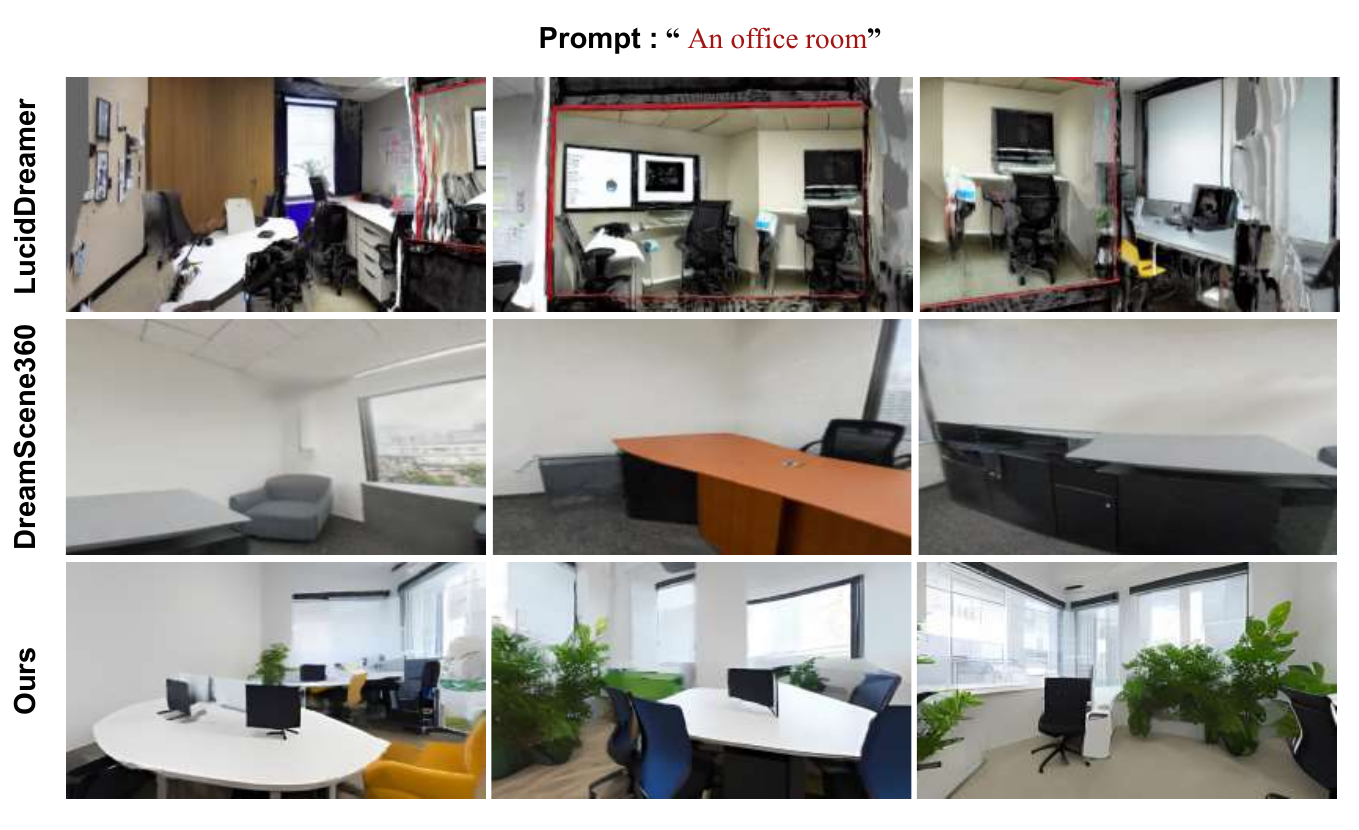} 
    \caption{Qualitative comparison of rendering results between our method and baseline approaches. Our method produces images with higher fidelity, better geometric details across different viewpoints.}
    \label{fig:render_comparison}
\end{figure}
\begin{table}
\centering
\caption{Ablation study results showing the impact of different layer combinations on model performance. layer1 represents layout reasoning layer, and layer2 represents self-optimization layer}
\label{tab:ablation_study_layer}
\begin{tblr}{
    cells = {c},
    vline{3} = {-}{},
    hline{1-2,6} = {-}{},
}
layer 1 & layer2 & clip-distance↓ & Q-Align↑ & BRISQUE↓  \\
$\times$  & $\times$  & 0.7313 & 0.0092 & 32.2608  \\
$\times$  & $\checkmark$  & 0.7252 & 0.0084 & 30.3939 \\
$\checkmark$  & $\times$  & 0.7318 & 0.0104 & 34.7502 \\
$\checkmark$  & $\checkmark$  & \textbf{0.7224} & \textbf{0.0105} & \textbf{29.2609} 
\end{tblr}
\end{table}

\begin{table}
\centering
\caption{Ablation study results showing the impact of different loss combinations on model performance.}
\label{tab:ablation_study_loss}
\begin{tblr}{
  cells = {c},
  vline{3} = {-}{},
  hline{1-2,6} = {-}{},
}
$\mathcal{L}_{geo}$ & $\mathcal{L}_{sem}$ & PSNR↑    & SSIM↑   & LPIPS↓  \\
$\times$  & $\times$  & 33.85 & 0.9630 & 0.0601 \\
$\times$  & $\checkmark$  & 33.77 & 0.9624 & 0.0612 \\
$\checkmark$  & $\times$  & 33.88 & 0.9631 & 0.0599 \\
$\checkmark$  & $\checkmark$  & \textbf{34.98} & \textbf{0.9707} & \textbf{0.0510} 
\end{tblr}
\end{table}
\subsection{Ablation Study}

To evaluate the impact of our design choices, we conducted ablation studies on both the two-layer architecture and the loss functions. As shown in Table \ref{tab:ablation_study_layer}, the combination of the layout reasoning and self-refinement layers yields the best overall results, with the former enhancing structural consistency and the latter refining perceptual quality. Likewise, Table \ref{tab:ablation_study_loss} shows that jointly using $\mathcal{L}{geo}$ and $\mathcal{L}{sem}$ leads to superior performance, as the two losses complement each other in ensuring geometric accuracy and semantic coherence.
\section{Conclusion}
\label{sec:conclusion}
In this work, we present PSGS, a novel two-stage framework for generating globally consistent 3D scenes from text descriptions. Our approach first introduces a two-layer panorama optimization architecture, combining layout reasoning for spatial semantics and iterative MLLM-based refinement for visual fidelity. The subsequent panorama sliding mechanism enables geometrically coherent point cloud initialization for 3D Gaussian Splatting, while our joint semantic-geometric consistency losses ensure structural accuracy and contextual alignment. Extensive validation demonstrates that PSGS significantly outperforms state-of-the-art methods in cross-view consistency and photorealism. This work provides an efficient, high-fidelity solution for immersive content creation. Future directions include modeling complex compositional text inputs and integrating physics-aware illumination for dynamic scenes.
{\small
\bibliographystyle{IEEEbib}
\bibliography{refs}
}

\end{document}